%% file: _main.tex
\input{_constants}
\arxiv 

\pdfoutput=1
\documentclass[10pt,twocolumn,letterpaper]{article}
\input{cvpr_header}
\begin{document}
\title{\paperTitle}
\author{\aauthor}
\maketitle

\input{00_abstract}
\input{01_intro}

\input{02_related}

\input{03_method}

\input{09_Experiment}

\input{10_conclusion}


{\small
\bibliographystyle{ieeenat_fullname}
\bibliography{11_references}
}

\end{document}

%% file: _constants.tex
\def\paperID{14446}
\def\confName{ICCV}
\def\confYear{2025}

\def\paperTitle{Beyond Human-prompting: Adaptive Prompt Tuning with Semantic Alignment for Anomaly Detection}
\def\aauthor{
Pi-Wei Chen\\
Silesian University of Technology\\
{\tt\small nf6111015@gs.ncku.edu.tw}

\and
Jerry Chun-Wei Lin\\
Silesian University of Technology\\
{\tt\small jerry.chun-wei.lin@polsl.pl}

\and
Wei-Han Chen \\
National Cheng Kung University\\
{\tt\small e34081155@ncku.edu.tw}

\and
Jia Ji \\
National Cheng Kung University\\
{\tt\small jijia20001118@gmail.com}

\and
Zih-Ching Chen \\
Nvidia AI Technology Center\\
{\tt\small virginiac@nvidia.com}

\and
Feng-Hao Yeh\\
National Cheng Kung University\\
{\tt\small yeh.feng.hao.110@gmail.com}

\and
Chao-Chun Chen\\
National Cheng Kung University\\
{\tt\small chencc@imis.ncku.edu.tw}
}

\newif\ifreview 
\newif\ifarxiv \newcommand{\arxiv}{\arxivtrue}
\newif\ifcamera 
\newif\ifrebuttal 

%% file: cvpr_header.tex
\ifreview \usepackage[review]{cvpr} \fi
\ifarxiv \usepackage[pagenumbers]{cvpr} \fi
\ifrebuttal \usepackage[rebuttal]{cvpr} \fi
\ifcamera \usepackage{cvpr} \fi

\input{_macros}  

\usepackage{xr-hyper}

\makeatletter
\newcommand*{\addFileDependency}[1]{
  \typeout{(#1)}
  \@addtofilelist{#1}
  \IfFileExists{#1}{}{\typeout{No file #1.}}
}

\makeatother

\definecolor{cvprblue}{rgb}{0.21,0.49,0.74}
\usepackage[pagebackref,breaklinks,colorlinks,citecolor=cvprblue]{hyperref}
\usepackage[capitalize]{cleveref}
\crefname{section}{Sec.}{Secs.}
\crefname{table}{Table}{Tables}
\crefname{figure}{Fig.}{Figs.}

%% file: _macros.tex
\usepackage{siunitx}
\usepackage{booktabs}
\usepackage{graphicx}	
\usepackage{amsmath}	
\usepackage{amssymb}	
\usepackage{booktabs}
\usepackage{times}
\usepackage{microtype}
\usepackage{epsfig}
\usepackage[table,xcdraw,dvipsnames]{xcolor}
\usepackage{caption}
\usepackage{float}
\usepackage{tabularray}
\usepackage{placeins}
\usepackage{color, colortbl}
\usepackage{stfloats}
\usepackage{enumitem}
\usepackage{tabularx}
\usepackage{xstring}
\usepackage{graphicx} 
\usepackage{multirow}
\usepackage{xspace}
\usepackage{url}
\usepackage{subcaption}
\usepackage{xcolor}
\usepackage[hang,flushmargin]{footmisc}

\ifcamera \usepackage[accsupp]{axessibility} \fi

% ...

\ifarxiv  \fi

\newcommand{\R}[1]{{%
    \textbf{%
        \ifstrequal{#1}{1}{\textcolor{red}{R#1}}{%
        \ifstrequal{#1}{2}{\textcolor{blue}{R#1}}{%
        \ifstrequal{#1}{3}{\textcolor{magenta}{R#1}}{%
        \ifstrequal{#1}{4}{\textcolor{teal}{R#1}}{%
                           \textcolor{cyan}{R#1}%
        }}}}%
    }%
}}

%% file: 00_abstract.tex
\begin{abstract}

Pre-trained Vision-Language Models (VLMs) have recently shown promise in detecting anomalies. However, previous approaches are fundamentally limited by their reliance on human-designed prompts and the lack of accessible anomaly samples, leading to significant gaps in context-specific anomaly understanding. In this paper, we propose \textbf{A}daptive \textbf{P}rompt \textbf{T}uning with semantic alignment for anomaly detection (APT), a groundbreaking prior knowledge-free, few-shot framework and overcomes the limitations of traditional prompt-based approaches. APT uses self-generated anomaly samples with noise perturbations to train learnable prompts that capture context-dependent anomalies in different scenarios. To prevent overfitting to synthetic noise, we propose a Self-Optimizing Meta-prompt Guiding Scheme (SMGS) that iteratively aligns the prompts with general anomaly semantics while incorporating diverse synthetic anomaly. Our system not only advances pixel-wise anomaly detection, but also achieves state-of-the-art performance on multiple benchmark datasets without requiring prior knowledge for prompt crafting, establishing a robust and versatile solution for real-world anomaly detection.

\end{abstract}

%% file: 01_intro.tex
\section{Introduction}
Anomaly detection often faces the challenge of obtaining annotations for anomalous samples. Therefore, previous researches have focused on unsupervised methods, leveraging large volumes of anomaly-free data to detect deviations during inference  \cite{xie2016unsupervised, fan2018abnormal, schlegl2019f,bergmann2020uninformed,chen2022utrad,you2022unified,lu2024hierarchical}. Recently, there has been a focus on scenarios with limited normal samples, where the scarcity of examples makes anomaly detection particularly challenging \cite{cohen2020patchcore,zhang2021regad,jeong2023winclip,li2024promptad,gu2024anomalygpt,zhou2023anomalyclip}. In these cases, prompt-based approaches that utilize pre-trained vision-language models (VLMs) have shown promising results \cite{jeong2023winclip,li2024promptad,gu2024anomalygpt,zhou2023anomalyclip}.
In VLM-based anomaly detection, the model utilize both textual and visual embeddings, capturing “normal” and “anomalous” states by aligning text prompts with visual cues within input images to detect anomaly. This alignment facilitates zero-shot and few-shot anomaly detection, enabling VLMs to adapt to new contexts without extensive labeled data.

\begin{figure}
\setlength{\abovecaptionskip}{0pt}
\setlength{\belowcaptionskip}{0pt}
\centering
\includegraphics[width=0.5\textwidth]{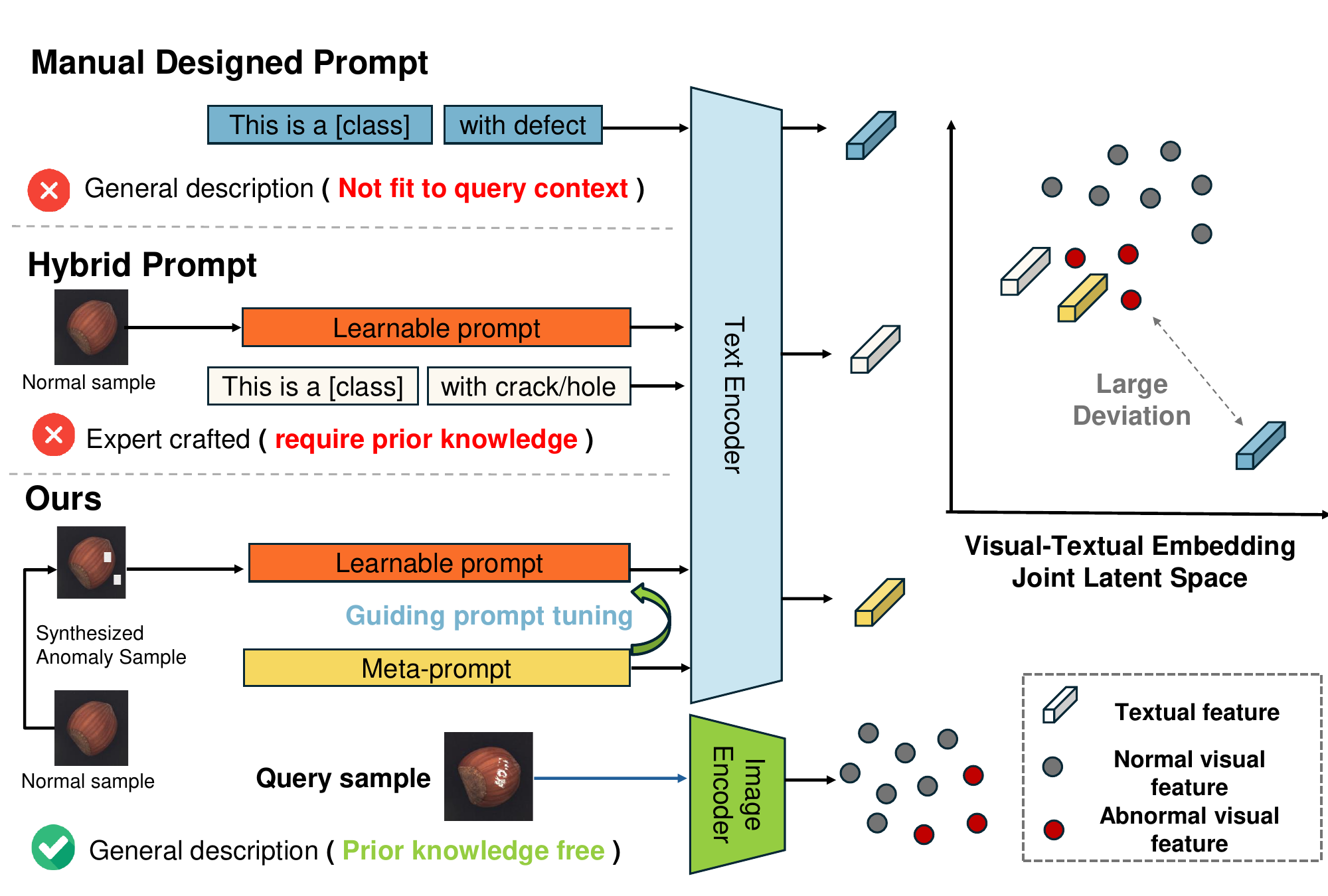}
\caption{ Comparison of previous methods and APT. The upper and middle section illustrates traditional approaches with manually crafted prompts, which lack context adaptability, require prior knowledge, and struggle with real-world anomaly alignment. The bottom section shows our data-driven, context-aware APT framework, which enhances prompt tuning to capture diverse anomaly patterns without prior knowledge, improving robustness across varied contexts.}
\label{The illustration of Fig1}
\end{figure}
In VLM-based anomaly detection, manual designed prompts like “this is a photo of [object] with a defect” indiscriminately  describe anomalies across all instances of the same object, overlooking the visual context, including details such as object properties, surface texture, and environmental conditions from query image\cite{jeong2023winclip}, thereby induce discrepancy between prompt and visual semantic. For instance, a scratch on a reflective surface differs from one on a matte background, affecting its representation within the feature space. As shown in the upper part of Figure~\ref{The illustration of Fig1}, where manual designed prompts can fail to adapt to the specific visual context, leading to mismatched embeddings that overlook the  unique characteristics of the query image.

To mitigate the limitations inherent in manually designed prompts, recent studies \cite{li2024promptad,zhou2023anomalyclip} have leveraged a small number of normal samples to perform prompt tuning \cite{zhou2022learning}, aiming to achieve better alignment with the context of query images. Nevertheless, without labeled anomaly examples, these tuned prompts predominantly reflect normal contextual characteristics, making it challenging to capture context-dependent anomaly semantics. Consequently, during inference, the prompts often lack sufficient contextual sensitivity to accurately identify anomalies across varying visual backgrounds, resulting in suboptimal detection performance \cite{zhou2023anomalyclip}. Although PromptAD \cite{li2024promptad} addresses this issue by introducing hybrid prompts—combining human-crafted and learnable components—the method still depends heavily on prior knowledge. This reliance limits its effectiveness when faced with previously unseen anomaly types, as illustrated in the middle section of Figure~\ref{The illustration of Fig1}.

An intuitive method to overcome this challenge involves generating synthetic anomalies by applying perturbations, such as Gaussian noise, to emulate deviations from normal patterns. However, a critical limitation of this approach is that artificially induced anomalies may not faithfully represent real-world anomaly semantics, leading to a misalignment between synthetic anomaly and real-world anomaly semantics  encountered in practical scenarios.

To address this issue, we propose \textbf{A}daptive \textbf{P}rompt \textbf{T}uning with Anomaly Semantic Alignment (APT), a novel framework designed to optimize learnable prompts for enhanced contextual sensitivity without requiring prior knowledge of specific anomalies. APT employs the Self-Optimizing Meta-Prompt Guiding Scheme (SMGS), in which a Meta-Prompt—initially a general manually designed prompt such as "this is a photo of an [object] with a defect"—serves as a dynamic anchor for guiding prompt optimization. SMGS ensures alignment with authentic anomaly semantics by continuously leveraging the Meta-Prompt to direct the optimization of learnable prompts, thereby capturing meaningful anomaly patterns while avoiding overfitting to artificial perturbations. After each iteration, the optimized learnable prompt replaces the previous Meta-Prompt, serving as a progressively refined anchor for subsequent tuning phases. This iterative refinement enhances the robustness and adaptability of the model across diverse anomalies and varying contextual environments.

To generate anomaly samples that better reflect context-dependent characteristics, we propose a \textbf{C}ontextual anomaly \textbf{F}eature \textbf{G}eneration (CFG) module. This module selectively applies perturbations only to regions of interest within the image, explicitly excluding irrelevant background areas. By doing so, CFG ensures the synthetic anomalies closely mirror context-specific anomalies that manifest exclusively on the target object.
Furthermore, to overcome localization limitations inherent in Vision-Language Models (VLMs) \cite{wang2023sclip, li2023clip}, we introduce a Locality-Aware Transformer (LAT) as the image encoder. LAT replaces the conventional attention mechanism with Locality Attention (LA), aligning feature map regions more precisely with their corresponding spatial locations in the image. This design significantly enhances the spatial alignment between textual prompts and targeted image regions, thereby improving the accuracy of pixel-level anomaly detection.

In summary, our proposed Adaptive Prompt Tuning (APT) framework facilitates learning optimal prompts without relying on prior human semantic knowledge. It effectively adapts to diverse contexts using only a few samples, thereby enabling precise pixel-wise anomaly detection, as illustrated in the lower portion of Figure~\ref{The illustration of Fig1}. The main contributions of this paper are summarized as follows:

\begin{itemize}
\item We develop Adaptive Prompt Tuning (APT), a novel framework that advances beyond human-designed prompts for Vision-Language Model (VLM)-based anomaly detection. APT automatically generates context-sensitive prompts tailored to detect anomalies across various objects without any reliance on manually crafted prompts or prior anomaly semantics.
\item We propose the Self-Optimizing Meta-Prompt Guiding Scheme (SMGS), a method that enables prompts to dynamically learn and capture context-dependent anomaly semantics from self-generated samples. SMGS iteratively refines the meta-prompt, surpassing static human-defined semantics and ensuring robustness against irrelevant noise, all without requiring prior knowledge of anomaly characteristics.
\item We develop the Contextual Anomaly Feature Generation (CFG) module, designed to synthesize anomaly samples with enhanced contextual relevance, thereby significantly improving the quality of samples used by SMGS. Additionally, we propose a Locality-Aware Transformer (LAT), a novel attention mechanism within the visual encoder, which effectively captures local spatial features and thus significantly boosts VLM-based anomaly detection performance.
\end{itemize}

\begin{figure*}
\setlength{\abovecaptionskip}{0pt}
\setlength{\belowcaptionskip}{0pt} 
\centering
\includegraphics[width=\linewidth]{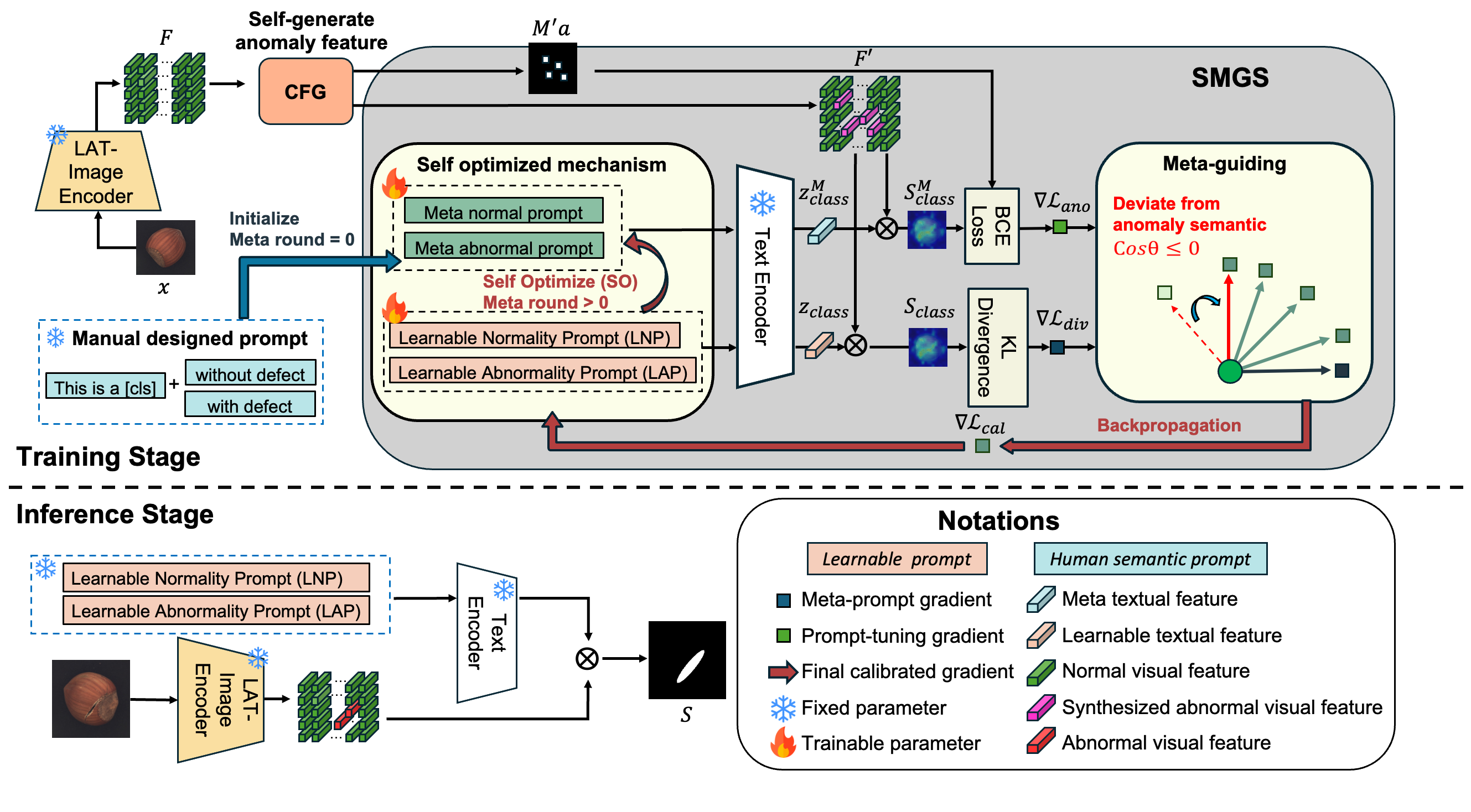}
   \caption{The overall architecture of the proposed Adaptive Prompt Tuning (APT) framework, featuring three main components: (1) Self-Optimizing Meta-Prompt Guiding Scheme (SMGS), which aligns learnable prompts with anomaly semantics through gradient calibration; (2) Contextual Feature Generation (CFG), which generates context-aware anomalies to enhance prompt adaptability; and (3) Locality-Aware Transformer (LAT), which focuses on locality-sensitive feature extraction. These components collaboratively optimize prompt alignment and improve anomaly detection robustness across diverse contexts. Upper part illustrate the prompt tuning process; lower part illustrate the inference stage for obtaining prompt-based score map.}
\label{The illustration of overall architecture}
\end{figure*}

%% file: 02_related.tex
\section{Related Work}
\label{sec:related}

\subsection{Vision-language model }
Vision-language models (VLMs) are sophisticated AI frameworks that integrate Computer Vision (CV) and Natural Language Processing (NLP) to align semantic representations from visual and textual data. Trained on extensive image-text paired datasets, these multimodal models leverage techniques such as contrastive learning \cite{chen2020simple} and masked language modeling to project visual features and textual embeddings into a unified semantic space. Notably, OPENCLIP \cite{schuhmann2022laionb}, a prominent open-source VLM, excels in generalization capabilities, enabling rapid and effective adaptation to diverse downstream tasks such as image classification, object detection \cite{kuo2022f}, and segmentation \cite{sun2024clip}, without task-specific training.
The adaptability of VLMs is further enhanced through prompt-based techniques, enabling controlled model behavior via carefully crafted textual prompts tailored to specific tasks. Recent advances by \cite{zhou2022learning} introduced context optimization strategies, employing learnable prompts tuned on few-shot examples. This approach significantly improves alignment between prompts and task-specific contexts, leading to superior performance compared to traditional static, human-designed prompts.

\subsection{Few-Shot Anomaly Detection}
Recent studies \cite{cao2023segment,jeong2023winclip,zhou2023anomalyclip,li2024promptad,gu2024anomalygpt} have leveraged multimodal frameworks based on vision-language models (VLMs) for few-shot and zero-shot anomaly detection, driven by textual prompting. These approaches exploit the semantic alignment between textual descriptions (e.g., "normal" vs. "anomalous") and visual data, enabling anomaly detection without extensive anomaly annotations and facilitating rapid domain adaptation.
WinCLIP \cite{jeong2023winclip} employs prompt ensembles heavily reliant on human-defined semantics, necessitating iterative experimentation to achieve optimal performance. Similarly, AnomalyGPT \cite{gu2024anomalygpt} utilizes manually crafted prompts that depend on prior knowledge and general state descriptions, often lacking the necessary context for detailed anomaly identification. Hybrid methods, such as AnomalyCLIP \cite{zhou2023anomalyclip}, blend learnable prefixes with manually designed suffixes, aiming to balance generalizability and specificity, yet remain constrained by predefined human semantics. PromptAD \cite{li2024promptad} adopts fully learnable prompts; however, it initializes learning from human-designed templates, thus inheriting inherent human limitations.
Collectively, these methods underscore the critical need for fully data-driven prompt optimization to enhance context-sensitive anomaly detection, thereby reducing dependency on human-crafted semantics and significantly improving detection robustness and precision.

%% file: 03_method.tex
\section{Proposed Framework}
\label{sec:method}

\subsection{Problem Definition}
In this paper, we address the few-shot anomaly detection task, where only a few normal samples \( D_{\text{train}} = \{(\mathbf{x}_i, y_i=0)\}_{i=1}^{k} \) are available for training, with each image \( \mathbf{x}_i \in \mathbb{R}^{H \times W \times C} \) representing a normal sample when \( y_i = 0 \). Anomalous samples, denoted as \( y_i = 1 \), are not provided in the training set, highlighting the challenge of detecting anomalies without prior exposure to abnormal data.

We introduce \textit{Learnable Normality Prompt} (LNP) and \textit{Learnable Abnormality Prompt} (LAP) as learnable textual embedding to distinguish normal and anomalous states.

During inference, an image encoder \(I(\cdot)\) maps an input image \(\mathbf{x}\) to a spatial feature map \(F \in \mathbb{R}^{s \times d}\), where \(s\) represents spatial locations and \(d\) denotes the embedding dimension:
\begin{equation}
I: \mathbb{R}^{H \times W \times C} \rightarrow \mathbb{R}^{s \times d}.
\end{equation}
Concurrently, a text encoder \(T(\cdot)\) maps the learnable prompts LNP and LAP into textual embeddings \(\mathbf{z}_n, \mathbf{z}_a \in \mathbb{R}^{d}\):
\begin{equation}
T: \mathbb{R}^{t \times d} \rightarrow \mathbb{R}^{d}.
\end{equation}
Anomaly scores \(S_n\) and \(S_a\) are computed as inner products between spatial features \(F\) and text embeddings \(\mathbf{z}_n, \mathbf{z}_a\):
\begin{equation}
S_n = F \cdot \mathbf{z}_n, \quad S_a = F \cdot \mathbf{z}_a.
\end{equation}
The final anomaly score map \(S\) is obtained by applying a softmax activation \(\sigma(\cdot)\) to the difference between abnormal and normal scores:
\begin{equation}
S = \sigma(S_a - S_n).
\end{equation}

The learning objective is to optimize LNP and LAP by minimizing the binary cross-entropy (BCE) loss \(\mathcal{L}_{\text{ano}}\) between the predicted anomaly map \(S\) and the ground truth mask \(M\):
\begin{equation}
\arg\min_{\text{LNP}, \text{LAP}} \mathcal{L}_{\text{ano}} = \text{BCE}(S, M).
\end{equation}

This optimization aligns predicted anomaly scores with the actual spatial anomaly distribution, enabling improved generalization to unseen anomalies.

\subsection{Overview of the Proposed Framework}
The overall architecture of our proposed \textbf{A}daptive \textbf{P}rompt \textbf{T}uning (APT) framework for anomaly detection is illustrated in Figure~\ref{The illustration of overall architecture}. At the core of this framework is the \textbf{Self-Optimizing Meta-Prompt Guiding Scheme (SMGS)} (Section~\ref{sec:Self-optimizing Meta-prompt Guiding Scheme}), designed to adaptively tune prompts to effectively capture context-aware anomaly semantics by leveraging synthetic anomalies generated from limited normal samples \( x \in D_{\text{train}} \).

To further enhance the effectiveness of SMGS, we develop the \textbf{Contextual Anomaly Feature Generation (CFG)} module (Section~\ref{CFG module}), which generates context-dependent synthetic anomalies. CFG selectively applies perturbations only to target pixels, thus ensuring the generated anomalies are aligned with the specific context of the target object. This targeted approach is crucial, as experiments indicate that employing random noise without context-aware constraints significantly reduces detection performance.

Additionally, to strengthen locality feature extraction, we propose the \textbf{Locality-Aware Transformer (LAT)} (Section~\ref{sec:Locality-aware attention}) as the visual encoder. LAT utilizes neighbor-cls-only attention, effectively capturing local spatial relationships and enhancing the precision of anomaly detection.
Together, these modules form a comprehensive and robust framework capable of efficiently adapting to new contexts without extensive labeled data.

\subsection{Learning Adaptive Context-sensitive Prompts}
\label{sec:Self-optimizing Meta-prompt Guiding Scheme}
To effectively detect context-dependent anomaly, we introduce LNP and LAP. These prompts are designed to capture the nuances of normal and abnormal features in different contexts of the query images, improving the adaptability and accuracy of anomaly detection.

Given the rarity and unpredictability of anomalies in real-world scenario, we synthesize anomaly samples by applying noise perturbations to a few-shot normal sample set, creating a controlled environment for training LNP and LAP. Specifically, the proposed CFG module (see Sec~\ref{CFG module}) generates contextually aligned anomalous sample \( \{(\mathbf{x}'_i, y_i=1, M'_a)\}_{i=1}^{k} \), where each \( M'_a \in \{0, 1\}^{H \times W} \) is a binary mask indicating synthetic anomaly locations. The normality mask \( M'_n \) is obtained as \( M'_n = 1 - M'_a \).

\subsubsection{Training Objective}  
To optimize the LNP and LAP, we define the binary cross-entropy (BCE) loss for both prompts in a unified form:

\begin{align}
\mathcal{L}_{\text{ano}}^{\textit{prompt}} =& - \frac{1}{H \times W} \sum_{j=1}^{H \times W}  \left[ M'_{\text{class}}(j) \log S'_{\textit{class}}(j) \right. \notag \\
& \left. + \left(1 - M'_{\textit{class}}(j)\right) \log \left(1 - S'_{\textit{class}}(j)\right) \right]
\end{align}

where \( \textit{class} \in \{a, n\} \), with \( a \) representing anomalies and \( n \) representing normality, and \( \textit{prompt} \in \{\text{LNP}, \text{LAP}\} \). Here, \( M'_{\text{class}} \) is the corresponding binary mask (anomaly \( M'_a \) or normal \( M'_n \)), and \( S'_{\textit{class}} \) is the score map produced by each prompt. The BCE loss functions \( \mathcal{L}_{\text{ano}}^{\text{LNP}} \) and \( \mathcal{L}_{\text{ano}}^{\text{LAP}} \) measure the discrepancy between the predicted score maps and the ground truth masks for normal and abnormal regions, respectively. By minimizing these losses, we effectively train LNP and LAP to produce score maps that accurately reflect the presence of normality or abnormality in the input images.

The total anomaly loss is then given by:
\begin{equation}
\mathcal{L}_{\text{ano}} = \mathcal{L}_{\text{ano}}^{\text{LNP}} + \mathcal{L}_{\text{ano}}^{\text{LAP}}
\end{equation}

\subsubsection{Meta-guiding}
To enhance the alignment of the learnable prompts with real-world anomaly semantics, we utilize meta-prompts and gradient calibration. We introduce the Meta Normality Prompt (MNP) and Meta Abnormality Prompt (MAP) as generalized representations of normality and abnormality. Importantly, the initial meta-prompts are generic prompt templates—such as "this is an object with/without defect"—that require no prior knowledge of specific anomaly types, allowing for broad applicability across various contexts. These meta-prompts act as guiding references during optimization to direct learnable prompts toward meaningful patterns and prevent overfitting to synthetic anomalies. Specifically, we compute their score maps \( S^M_{\textit{class}} = \sigma\left( F' \cdot \mathbf{z}^M_{\textit{class}} \right) \), where \( \mathbf{z}^M_{\textit{class}} \) are the embeddings of MNP and MAP.\\
The divergence loss between the learnable prompts and meta-prompts is:
\begin{equation}
\label{eq:div_loss}
\mathcal{L}_{\text{div}}^{\textit{prompt}} = \text{KL}\left( S^M_{\textit{class}} \parallel S'_{\textit{class}} \right)
\end{equation}

with the total divergence loss:

\begin{equation}
\label{eq:total_div_loss}
\mathcal{L}_{\text{div}} = \mathcal{L}_{\text{div}}^{\text{LNP}} + \mathcal{L}_{\text{div}}^{\text{LAP}}
\end{equation}
\textbf{Gradient Calibration:}
During optimization, conflicts can arise between the gradients from the anomaly loss \( \mathcal{L}_{\text{ano}} \), which focuses on detecting synthetic anomalies, and the divergence loss \( \mathcal{L}_{\text{div}} \), which promotes alignment with meta-prompts representing general anomaly semantics. Such conflicts may lead the learnable prompts (LNP and LAP) to overfit to synthetic anomalies rather than capturing robust, generalizable features.

To resolve these conflicts, we apply a gradient calibration approach that adjusts \( \nabla \mathcal{L}_{\text{ano}} \) based on its alignment with \( \nabla \mathcal{L}_{\text{div}} \). Specifically, when the cosine similarity \( C = \cos\left( \nabla \mathcal{L}_{\text{ano}}, \nabla \mathcal{L}_{\text{div}} \right) \) is negative, \(  C \cdot \nabla\mathcal{L}_{\text{div}}\) signifies a direction that is contrary to \(\nabla\mathcal{L}_{\text{div}}\). Therefore, to realign the optimization toward general anomaly semantics, we subtract  \( \lambda \cdot C \cdot \nabla\mathcal{L}_{\text{div}}\) from \( \nabla \mathcal{L}_{\text{ano}} \):

\begin{equation}
\label{eq:grad_cal}
\nabla \mathcal{L}_{\text{cal}} =
\begin{cases}
\nabla \mathcal{L}_{\text{ano}} - \lambda \cdot C \cdot \nabla \mathcal{L}_{\text{div}}, & \text{if } C < 0 \\
\nabla \mathcal{L}_{\text{ano}}, & \text{otherwise.}
\end{cases}
\end{equation}

Here, \( \lambda \) controls the adjustment strength, modulating the influence of \( \nabla \mathcal{L}_{\text{div}} \) on the calibrated gradient. Intuitively, \( C \) represents the alignment between the gradients: a negative \( C \) indicates misalignment, suggesting that focusing solely on synthetic features may drive the prompts away from general anomaly semantics. 

\subsubsection{Self-Optimization Mechanism} 
By updating the meta-prompts with the newly optimized learnable prompts at the end of each meta-round, we enable the guiding signals to become increasingly tailored to the specific anomaly characteristics in the query image. Initially, the meta-prompts are general, manually crafted prompts that capture only broad anomaly concepts. In each meta-round \( T \), the model undergoes several regular training rounds (or epochs) to refine the LNP and LAP prompts for more specific anomaly detection in the query context. After completing the training rounds in each meta-round, we replace the meta-prompts with the optimized learnable prompts, preparing them to serve as meta-prompts for the next meta-round \( T+1 \), as follows:

\begin{equation} \label{eq}
\text{MNP}^{T+1} \leftarrow \text{LNP}^{T}, \quad \text{MAP}^{T+1} \leftarrow \text{LAP}^{T}
\end{equation}

This iterative updating of the meta-prompts after each meta-round progressively improves the model’s robustness and adaptability to various anomaly types and contexts, thereby enhancing the guidance provided to LNP and LAP in successive rounds.

\subsection{Contextual-Anomaly Feature Generation Module}
\label{CFG module}
Instead of applied gaussian randomly across whole image of few-shot sample, we proposed CFG module to applied noise only on target item within the image for contextual-aligned anomaly sample. 

The CFG module processes the feature map of few-shot samples $F$ as input. To isolate the target object, a binary mask \( \mathbf{M}_{\text{obj}} \) is generated in a \textbf{Target Focus (TF)} mechanism. TF utilizes prompts such as ``This is a photo of \{object\}.'' to identify the target item mask in the feature map The mask is derived by computing the inner product between the text embedding \( \mathbf{z}_{\text{obj}} \) and the feature map \( F \), followed by applying a threshold of 0.5 to the resulting score map.:

\[
\mathbf{M}_{\text{obj}}(i,j) = 
\begin{cases} 
1 & \text{if } (F(i,j) \cdot \mathbf{z}_{\text{obj}}) > 0.5 \\
0 & \text{otherwise}
\end{cases}
\]

Gaussian noise \( \mathcal{N}(0, \sigma^2) \) is then applied to locations identified by \( \mathbf{M}_{\text{obj}} \), producing synthetic anomalies:

\[
F'(i,j) = 
\begin{cases} 
F(i,j) + \mathcal{N}(0, \sigma^2) & \text{if } \mathbf{M}_{\text{obj}}(i,j) = 1 \\
F(i,j) & \text{otherwise}
\end{cases}
\]

These contextually-aligned synthetic samples \( \{(F'_i, y_i=1, M'_a)\}_{i=1}^{k} \) are used for tuning \textbf{LNP} and \textbf{LAP} in SMGS, enhancing its ability to detect anomalies with greater contextual precision.

\subsection{Locality-aware Transformer}
\label{sec:Locality-aware attention}
The vanilla attention mechanism in Vision Transformer (ViT) is not well-suited for pixel-level tasks due to a \textbf{feature misalignment} issue \cite{wang2023sclip, li2023clip}. As shown in \cite{dosovitskiy2020image}, this misalignment arises because the original attention mechanism allows each patch (token) to attend to distant unrelated patches, causing local features to blend with background details and creating a mismatch between input and output features.

To solve this problem, we introduce the \textbf{Locality-Aware Transformer (LAT)}, which is specifically designed for tasks that require spatial precision. The LAT uses a novel Locality Attention (LA) mechanism that restricts each token to pay attention only to its neighboring tokens, thereby preserving important local spatial details and preventing irrelevant background blending.

To maintain stability and avoid cascading changes across layers, we employ a dual-path design inspired by \cite{li2022exploring}. This structure retains the original global attention path alongside the new locality-restricted path, allowing the model to capture both local and global context without disrupting layer dependencies.

In the LAT, we apply a $k$-neighbor mask \( \mathbf{M}_{\text{LA}} \) in the attention map, ensuring each token attends only to its closest neighbors:

\begin{equation}
    \text{Attention}_{\text{LA}}(\mathbf{Q}, \mathbf{K}, \mathbf{V}) = \text{softmax} \left( \frac{\mathbf{Q} \mathbf{K}^T + \mathbf{M}_{\text{LAA}}}{\sqrt{d_k}} \right) \mathbf{V},
\end{equation}
where \( \mathbf{M}_{\text{LA}} \) is defined to limit each token’s attention as follows:

\begin{equation}
    M_{(i,j),(m,n)} =
\begin{cases}
0 & \text{if } \sqrt{(i - m)^2 + (j - n)^2} \leq k \\
-\infty & \text{otherwise}
\end{cases}
\end{equation}

Here, \( (i, j) \) and \( (m, n) \) represent coordinates on a 2D grid of patches, with \( k \) specifying the neighborhood size. The LAT’s dual-path and locality-focused design ensures that spatially precise features are preserved while also capturing global context, making it well-suited for fine-grained tasks.

%% file: 09_Experiment.tex
\section{Experiments}
\label{sec:experiment}
In this paper, we compare APT with recent VLM-based anomaly detection methods \cite{jeong2023winclip,zhou2023anomalyclip,li2024promptad,gu2024anomalygpt} over five datasets to evaluate the generalization performance. Few-shot method comparisons \cite{park2019semantic,defard2021padim,roth2022towards} and class-level results are included in the Appendix, as our approach targets pixel-wise anomaly detection. In addition, an ablation study is performed on each APT component to verify its individual contribution.
\\
\textbf{Dataset}
In this paper, we conduct experiments with the datasets MVTec \cite{bergmann2019mvtec}, VisA \cite{zou2022spot}, MPDD \cite{MPDD2021}, SDD \cite{Tabernik2019JIM}, which originate from the industrial domain, and we also use a medical dataset ColonDB \cite{bernal2012towards} for domain generalization evaluation.
The above datasets all provided with pixel-level ground-truth mask for anomaly segmentation.\\
\textbf{Evaluation metrics}
In our evaluation of anomaly detection, we use the Area Under the Receiver Operating Characteristic curve (AUROC) as the primary metric. The AUROC is ideal for this task as it captures the model's ability to distinguish between true positives and false positives at different thresholds. Considering the full range of possible thresholds, AUROC provides a robust measure of the model's performance and is therefore well suited for evaluating its effectiveness in detecting anomalies.
\\
\textbf{Implementation Details}
The proposed APT framework is fully compatible with the Vision-Guided (VG) mechanism used in previous prompt-based anomaly detection methods such as WinCLIP~\cite{jeong2023winclip} and PromptAD~\cite{li2024promptad}. Following the standard setup described in AnomalyCLIP~\cite{zhou2023anomalyclip}, we extract intermediate spatial features (excluding the CLS token) from two specific layers of the visual encoder during training, forming a normal visual memory bank (denoted as \( R \)). In the testing phase, we obtain the vision-guided anomaly score map \( S_v \) by calculating the cosine similarity between the spatial feature maps \( F \in \mathbb{R}^{s \times d} \) of the query image and the stored visual memory \( R \), subsequently averaging scores from these two layers.
The final score map with VG $S_{VG}$ is calulated as $S_{VG}=S+S_v$.

 Each model is optimized for 100 epochs with 5 meta-rounds to iteratively refine the meta-prompts. We set the meta-guiding weight \(\lambda = 1\) to ensure stable optimization.
All experiments were conducted using a single NVIDIA 2080Ti GPU, employing the Adam optimizer with an initial learning rate of 1e-4, decayed by cosine annealing. Visual and textual features are \(\ell_2\)-normalized before computing similarity scores, significantly enhancing numerical stability and consistency.

\subsection{Overall Performance}
Our APT framework sets a new benchmark in pixel-wise anomaly detection (see Table~\ref{tab:SOTA}), consistently outperforming state-of-the-art VLM-based methods—including WinCLIP, PromptAD, and AnomalyGPT—across all datasets and settings, even without prior knowledge for crafting prompts. Notably, in the 1-shot setting without VG, APT achieves the highest AUROC scores: 92.32$\pm$0.99 on MVTec (an improvement of approximately +1.0 over PromptAD and +10.92 over WinCLIP); 93.43$\pm$0.15 on VisA (+1.09 over PromptAD); 95.11$\pm$0.82 on MPDD (+0.34 over PromptAD); 95.73$\pm$0.54 on SDD (+3.72 over PromptAD); and 72.46$\pm$0.22 on CVC, significantly surpassing both PromptAD and WinCLIP. Even with the VG mechanism enabled, APT continues to lead—achieving 95.32$\pm$0.37 on MVTec in the 1-shot setting, thereby outperforming AnomalyGPT and PromptAD. In both the 2-shot and 4-shot settings, APT maintains its superior performance across all datasets. Compared to zero-shot methods (see Table~\ref{tab:AD Comparison with Zero-shot VLM-Method}), APT significantly boosts accuracy with only a few shots, confirming the advantage of utilizing minimal anomaly-free samples for prompt refinement. In general, these results affirm that our APT framework not only sets a new standard in anomaly detection, but also validates the efficacy and robustness of our approach, even without prior prompt crafting.


\begin{table*}[]
\caption{Qualitative comparisons with the SOTA VLM-based methods for pixel-wise detection of anomalies. The best AUROC values are highlighted in \textbf{Bold}.}
\label{tab:SOTA}
\resizebox{1\textwidth}{!}{
\begin{tabular}{|llllllllllll|}
\hline
\multicolumn{12}{|c|}{\textbf{Pixel-wise Anomaly Detection Comparison Table}} \\ \hline
 & \multicolumn{1}{l|}{Method} & \multicolumn{2}{c|}{MVTec} & \multicolumn{2}{c|}{VisA} & \multicolumn{2}{c|}{MPDD} & \multicolumn{2}{c|}{SDD} & \multicolumn{2}{c|}{CVC} \\ \hline
 & \multicolumn{1}{l|}{} & w/o VG & \multicolumn{1}{l|}{w/ VG} & w/o VG & \multicolumn{1}{l|}{w/ VG} & w/o VG & \multicolumn{1}{l|}{w/ VG} & w/o VG & \multicolumn{1}{l|}{w/ VG} & w/o VG & w/ VG \\ \hline
\multicolumn{1}{|l|}{\multirow{4}{*}{1-shot}} 
& \multicolumn{1}{l|}{WinCLIP \cite{jeong2023winclip}} 
& 81.40 $\pm$ 0.87 & \multicolumn{1}{l|}{86.20 $\pm$ 0.55} 
& 85.67 $\pm$ 0.66 & \multicolumn{1}{l|}{89.96 $\pm$ 0.44} 
& 89.56 $\pm$ 0.71 & \multicolumn{1}{l|}{92.83 $\pm$ 0.33} 
& 68.02 $\pm$ 0.97 & \multicolumn{1}{l|}{70.91 $\pm$ 0.25} 
& 63.24 $\pm$ 0.42 & 62.32 $\pm$ 0.88 \\ \cline{2-12}
\multicolumn{1}{|l|}{} 
& \multicolumn{1}{l|}{AnomalyGPT \cite{gu2024anomalygpt}} 
& - & \multicolumn{1}{l|}{95.30 $\pm$ 0.11} 
& - & \multicolumn{1}{l|}{\textbf{96.20 $\pm$ 0.99}} 
& - & \multicolumn{1}{l|}{-} 
& - & \multicolumn{1}{l|}{-} 
& - & - \\ \cline{2-12}
\multicolumn{1}{|l|}{} 
& \multicolumn{1}{l|}{PromptAD \cite{li2024promptad}} 
& 91.32 $\pm$ 0.52 & \multicolumn{1}{l|}{95.00 $\pm$ 0.17} 
& 92.34 $\pm$ 0.64 & \multicolumn{1}{l|}{96.12 $\pm$ 0.23} 
& 94.77 $\pm$ 0.76 & \multicolumn{1}{l|}{95.53 $\pm$ 0.89} 
& 92.01 $\pm$ 0.47 & \multicolumn{1}{l|}{93.40 $\pm$ 0.30} 
& 62.46 $\pm$ 0.93 & 54.43 $\pm$ 0.12 \\ \cline{2-12}
\multicolumn{1}{|l|}{} 
& \multicolumn{1}{l|}{\textcolor{blue}{Ours}} 
& \textbf{92.32 $\pm$ 0.99} & \multicolumn{1}{l|}{\textbf{95.32 $\pm$ 0.37}} 
& \textbf{93.43 $\pm$ 0.15} & \multicolumn{1}{l|}{96.09 $\pm$ 0.67} 
& \textbf{95.11 $\pm$ 0.82} & \multicolumn{1}{l|}{\textbf{95.82 $\pm$ 0.45}} 
& \textbf{95.73 $\pm$ 0.54} & \multicolumn{1}{l|}{\textbf{95.47 $\pm$ 0.88}} 
& \textbf{72.46 $\pm$ 0.22} & \textbf{64.40 $\pm$ 0.36} \\ \hline
\multicolumn{1}{|l|}{\multirow{4}{*}{2-shot}} 
& \multicolumn{1}{l|}{WinCLIP \cite{jeong2023winclip}} 
& 81.40 $\pm$ 0.44 & \multicolumn{1}{l|}{89.65 $\pm$ 0.73} 
& 85.67 $\pm$ 0.98 & \multicolumn{1}{l|}{90.26 $\pm$ 0.12} 
& 89.56 $\pm$ 0.34 & \multicolumn{1}{l|}{92.78 $\pm$ 0.87} 
& 68.02 $\pm$ 0.58 & \multicolumn{1}{l|}{70.02 $\pm$ 0.93} 
& - & - \\ \cline{2-12}
\multicolumn{1}{|l|}{} 
& \multicolumn{1}{l|}{AnomalyGPT \cite{gu2024anomalygpt}} 
& - & \multicolumn{1}{l|}{95.60 $\pm$ 0.45} 
& - & \multicolumn{1}{l|}{96.40 $\pm$ 0.72} 
& - & \multicolumn{1}{l|}{-} 
& - & \multicolumn{1}{l|}{-} 
& - & - \\ \cline{2-12}
\multicolumn{1}{|l|}{} 
& \multicolumn{1}{l|}{PromptAD \cite{li2024promptad}} 
& 91.34 $\pm$ 0.63 & \multicolumn{1}{l|}{96.03 $\pm$ 0.14} 
& 92.52 $\pm$ 0.49 & \multicolumn{1}{l|}{96.43 $\pm$ 0.66} 
& 94.30 $\pm$ 0.21 & \multicolumn{1}{l|}{96.00 $\pm$ 0.97} 
& 91.00 $\pm$ 0.78 & \multicolumn{1}{l|}{90.80 $\pm$ 0.13} 
& - & - \\ \cline{2-12}
\multicolumn{1}{|l|}{} 
& \multicolumn{1}{l|}{\textcolor{blue}{Ours}} 
& \textbf{92.12 $\pm$ 0.26} & \multicolumn{1}{l|}{\textbf{97.00 $\pm$ 0.34}} 
& \textbf{93.61 $\pm$ 0.73} & \multicolumn{1}{l|}{\textbf{96.89 $\pm$ 0.45}} 
& \textbf{95.28 $\pm$ 0.68} & \multicolumn{1}{l|}{\textbf{96.32 $\pm$ 0.59}} 
& \textbf{96.68 $\pm$ 0.37} & \multicolumn{1}{l|}{\textbf{92.43 $\pm$ 0.86}} 
& - & - \\ \hline
\multicolumn{1}{|l|}{\multirow{4}{*}{4-shot}} 
& \multicolumn{1}{l|}{WinCLIP \cite{jeong2023winclip}} 
& 81.43 $\pm$ 0.88 & \multicolumn{1}{l|}{90.63 $\pm$ 0.31} 
& 85.67 $\pm$ 0.65 & \multicolumn{1}{l|}{91.30 $\pm$ 0.99} 
& 89.56 $\pm$ 0.14 & \multicolumn{1}{l|}{93.26 $\pm$ 0.72} 
& 68.02 $\pm$ 0.55 & \multicolumn{1}{l|}{71.10 $\pm$ 0.49} 
& - & - \\ \cline{2-12}
\multicolumn{1}{|l|}{} 
& \multicolumn{1}{l|}{AnomalyGPT \cite{gu2024anomalygpt}} 
& - & \multicolumn{1}{l|}{96.25 $\pm$ 0.28} 
& - & \multicolumn{1}{l|}{96.70 $\pm$ 0.47} 
& - & \multicolumn{1}{l|}{-} 
& - & \multicolumn{1}{l|}{-} 
& - & - \\ \cline{2-12}
\multicolumn{1}{|l|}{} 
& \multicolumn{1}{l|}{PromptAD \cite{li2024promptad}} 
& 91.63 $\pm$ 0.57 & \multicolumn{1}{l|}{96.34 $\pm$ 0.38} 
& 92.33 $\pm$ 0.81 & \multicolumn{1}{l|}{96.90 $\pm$ 0.69} 
& 93.53 $\pm$ 0.62 & \multicolumn{1}{l|}{96.52 $\pm$ 0.95} 
& 90.64 $\pm$ 0.14 & \multicolumn{1}{l|}{94.23 $\pm$ 0.22} 
& - & - \\ \cline{2-12}
\multicolumn{1}{|l|}{} 
& \multicolumn{1}{l|}{\textcolor{blue}{Ours}} 
& \textbf{92.16 $\pm$ 0.90} & \multicolumn{1}{l|}{\textbf{97.10 $\pm$ 0.37}} 
& \textbf{94.68 $\pm$ 0.56} & \multicolumn{1}{l|}{\textbf{97.10 $\pm$ 0.68}} 
& \textbf{95.14 $\pm$ 0.73} & \multicolumn{1}{l|}{\textbf{96.90 $\pm$ 0.92}} 
& \textbf{96.33 $\pm$ 0.14} & \multicolumn{1}{l|}{\textbf{95.30 $\pm$ 0.63}} 
& - & - \\ \hline
\end{tabular}
}
\end{table*}

\begin{table}[]
\centering
\caption{Pixel-wise AD comparison with Zero-shot VLM-Method. Best AUROCs are highlighted in \textbf{Bold}. }
\label{tab:AD Comparison with Zero-shot VLM-Method}
\resizebox{0.35\textwidth}{!}{
\begin{tabular}{|lll|}

\hline
\multicolumn{3}{|l|}{AD Comparison with Zero-shot VLM-method}                               \\ \hline
\multicolumn{1}{|l|}{Method}       & \multicolumn{1}{c|}{MVTec} & \multicolumn{1}{c|}{VisA} \\ \hline
\multicolumn{1}{|l|}{AnomalyCLIP \cite{zhou2023anomalyclip}}  & \multicolumn{1}{c|}{91.16}  & \multicolumn{1}{c|}{95.51}                     \\
\multicolumn{1}{|l|}{Ours (1-shot)} & \multicolumn{1}{c|}{\textbf{95.32}} & \multicolumn{1}{c|}{\textbf{96.09}}                    \\ \hline
\end{tabular}
}
\end{table}
\subsection{Ablation Study}

\begin{table}[]
\centering
\caption{Ablation Study}
\label{tab:Ablation study}
\resizebox{0.37\textwidth}{!}{
\begin{tabular}{|lllllll|}
\hline
\multicolumn{7}{|c|}{Ablation Study Results}                                                                                                                                                                                                                                                                                                     \\ \hline
\multicolumn{2}{|c|}{SMGS}                                                                            & \multicolumn{2}{c|}{CFG}                                                                             & \multicolumn{1}{l|}{\multirow{2}{*}{LA}}                    & \multicolumn{1}{c}{\multirow{2}{*}{MVtec}} & \multirow{2}{*}{VisA} \\ \cline{1-4}
SO                                     & \multicolumn{1}{l|}{MG}                                     & GN                                     & \multicolumn{1}{l|}{TF}                                     & \multicolumn{1}{l|}{}                                       & \multicolumn{1}{c}{}                       &                       \\ \hline
{\color{red}$\times$}                  & \multicolumn{1}{l|}{{\color{red}$\times$}}                  & {\color{red}$\times$}                  & \multicolumn{1}{l|}{{\color{red}$\times$}}                  & \multicolumn{1}{l|}{{\color{red}$\times$}}                  & 85.00                                       & 86.43                    \\
{\color{red}$\times$}                  & \multicolumn{1}{l|}{{\color{red}$\times$}}                  & {\color{green}$\checkmark$}            & \multicolumn{1}{l|}{{\color{red}$\times$}}                  & \multicolumn{1}{l|}{{\color{red}$\times$}}                  & 84.52                                       & 88.53                  \\
{\color{green}$\checkmark$}            & \multicolumn{1}{l|}{{\color{green}$\checkmark$}}            & {\color{green}$\checkmark$}            & \multicolumn{1}{l|}{{\color{red}$\times$}}                  & \multicolumn{1}{l|}{{\color{red}$\times$}}                  & 90.95                                       & 91.63                  \\
{\color{green}$\checkmark$}            & \multicolumn{1}{l|}{{\color{green}$\checkmark$}}            & {\color{green}$\checkmark$}            & \multicolumn{1}{l|}{{\color{green}$\checkmark$}}            & \multicolumn{1}{l|}{{\color{red}$\times$}}                  & 91.94                                       & 93.00                  \\
{\color{red}$\times$}                  & \multicolumn{1}{l|}{{\color{green}$\checkmark$}}            & {\color{green}$\checkmark$}            & \multicolumn{1}{l|}{{\color{green}$\checkmark$}}            & \multicolumn{1}{l|}{{\color{red}$\times$}}                  & 91.82                                       & 92.31                  \\
{\color{green}$\checkmark$}            & \multicolumn{1}{l|}{{\color{green}$\checkmark$}}            & {\color{green}$\checkmark$}            & \multicolumn{1}{l|}{{\color{green}$\checkmark$}}            & \multicolumn{1}{l|}{{\color{green}$\checkmark$}}            & 92.32                                       & 93.43                  \\ \hline
\end{tabular}}
\end{table}
To evaluate the contribution of each component in APT, we conducted an ablation study on the MVTec and VisA datasets, as summarized in Table~\ref{tab:Ablation study}. The analysis focuses on the SMGS, the CFG module, and the LAT.
The baseline (first row) represents the manually crafted prompt setting without any prompt tuning—i.e., with SMGS, CFG (both GN and TF), and LA all disabled—yielding 85.00\% on MVTec and 86.43\% on VisA.

\textbf{SMGS}: When prompt tuning is attempted without SMGS, the system overfits to irrelevant noise. This is evident in the second row, where enabling only Gaussian Noise (GN) from the CFG module (while keeping SMGS disabled) results in a slight drop in MVTec accuracy (from 85.00\% to 84.52\%) even though VisA accuracy increases to 88.53\%. In contrast, the introduction of SMGS (i.e., enabling both SO and MG in row three) significantly boosts performance to 90.95\% on MVTec and 91.63\% on VisA. This improvement highlights that SMGS effectively aligns prompt tuning with the overall semantics of anomalies, thereby preventing overfitting by guiding the meta-prompt to focus on meaningful anomaly patterns.

\textbf{CFG Module}: The CFG module is designed to facilitate context-dependent prompt tuning by generating anomaly features. While using only Gaussian Noise (GN) without the Target Focus (TF) results in suboptimal performance (as seen in row two), combining GN with TF under the guidance of SMGS (row four) improves accuracy to 91.94\% on MVTec and 93.00\% on VisA. This shows that TF introduces perturbations in semantically significant regions, thereby providing anomalies that more closely mirror real-world conditions. A comparison with the configuration without the self-optimization mechanism (row five, where SO is disabled but MG, GN, and TF remain enabled) shows a slight performance drop (91.82\% on MVTec and 92.31\% on VisA), confirming that the self-optimization (SO) aspect of SMGS refines the meta-prompt by capturing more context-specific anomaly features.

\textbf{LAT}: Finally, we examine the impact of the Locality-Aware Transformer (LAT). As shown in Table~\ref{tab:Ablation study}, adding LAT (row six) enhances performance from 91.94\% (with vanilla attention, row four) to 92.32\% for MVTec, and from 93.00\% to 93.43\% for VisA. This improvement clearly demonstrates the advantage of LAT, emphasizing that locality-aware attention, which enforces precise spatial correspondence between learned features and specific image regions, is essential for achieving robust, precise, pixel-level anomaly localization.

In summary, this ablation study confirms that each component of the APT framework, SMGS (with MG and SO), the CFG module (with GN and TF) and the LAT, plays a vital role in improving anomaly detection performance. While prompt tuning without SMGS leads to overfitting, the meta-guiding mechanism in SMGS, enhanced by self-optimization, aligns the learned prompts with the semantic structure of anomalies. The CFG module’s targeted perturbations provide contextually relevant training samples, and the LA Transformer ensures precise spatial alignment, collectively enabling APT to achieve state-of-the-art results in pixel-wise anomaly detection.

\section{Discussion}
\subsection{The Effectiveness of Self-Optimize Technique in SMGS}
\begin{figure}
\setlength{\abovecaptionskip}{0pt}
\setlength{\belowcaptionskip}{0pt}
\centering
\includegraphics[width=\linewidth]{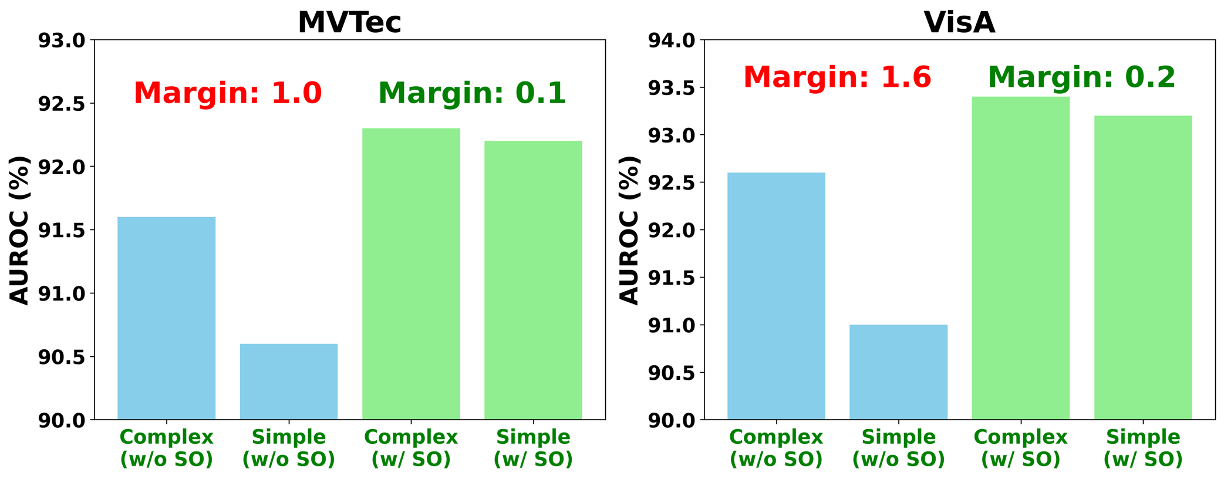}
\caption{Comparison of AUROC performance between complex and simple prompts on MVTec and VisA datasets with and without the Self-Optimize (SO) mechanism. Without SO, the performance margins are larger (1.0 for MVTec, 1.6 for VisA), indicating a reliance on complex prompts. With SO, the margins shrink considerably (0.1 for MVTec, 0.2 for VisA), showing the ability of SO to improve robustness even for simple prompts.}
\label{The Robustness of Self-optimize Mechanism}
\end{figure}

To evaluate the impact of the Self-Optimize (SO) mechanism on reducing the reliance on prior knowledge in prompt crafting, we evaluated complex and simple meta-prompts in the MVTec and VisA datasets (see Figure~\ref{The Robustness of Self-optimize Mechanism}). Without SO, complex prompts—crafted with extensive prior knowledge— performed slightly better, with AUROC values of 91.6\% versus 90.6\% for MVTec and 92.6\% versus 91.0\% for VisA. This resulted in performance margins of 1.0 and 1.6 between complex and simple prompts respectively, indicating a dependence on complex prompts for robust results in the absence of SO.
In contrast, enabling SO led to a significant reduction in the differences in performance in both datasets: For MVTec, the margin shrank to only 0.1 and for VisA to 0.2. This minimal gap shows that SO is able to iteratively refine even simple, knowledge-free prompts to achieve a competitive strength close to that of complex prompts. SO thus effectively generalizes the effectiveness of prompts without requiring complicated prior knowledge, and thus can be adapted to different contexts of anomaly detection with minimal prompt engineering effort.

\subsection{Comparative Analysis of Learnable Prompts on Synthesized vs. Real Anomalies}
\begin{figure}
\setlength{\abovecaptionskip}{0pt}
\setlength{\belowcaptionskip}{0pt}
\centering
\includegraphics[width=0.75\linewidth]{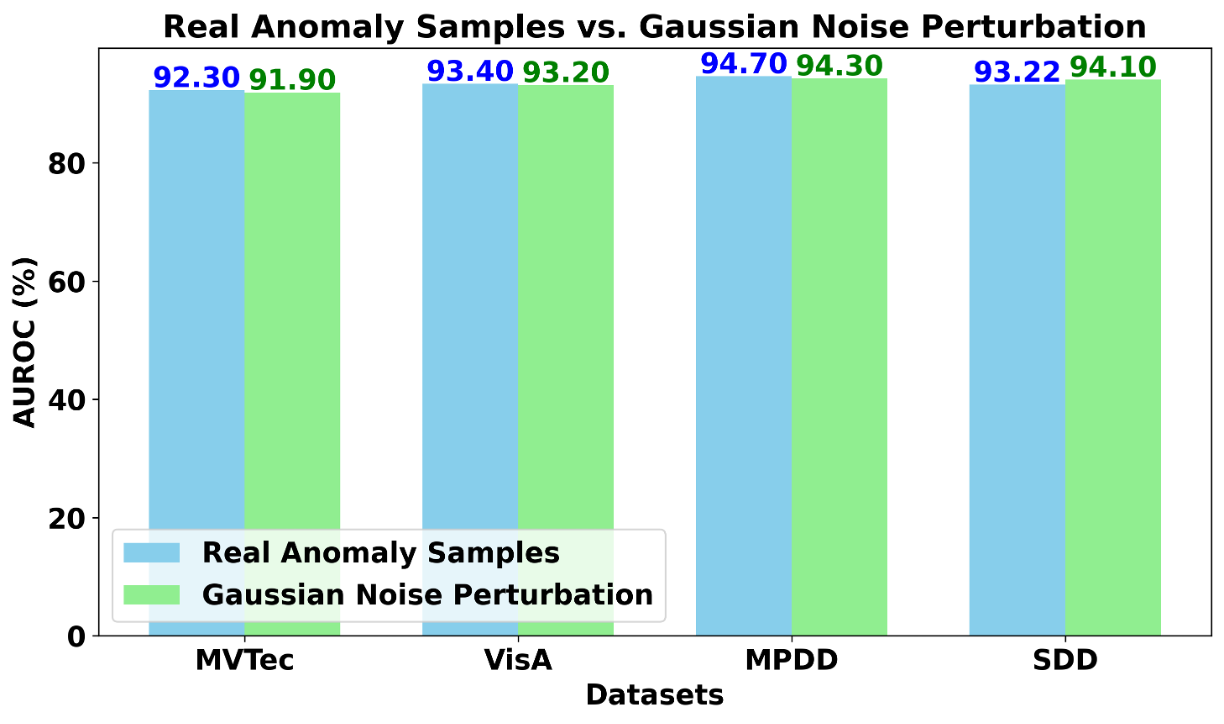}
\caption{Performance comparison between the use of real anomaly samples (blue) and Gaussian noise (green) in the MVTec, VisA, MPDD and SDD datasets. Gaussian noise perturbation maintains competitive performance, with an improvement over SDD, highlighting its utility as a flexible, self-generated anomaly strategy.}
\label{GA_eff}
\end{figure}

To explore the effectiveness of the Gaussian noise perturbation as a self-generated anomaly strategy, we performed a comparison with a few-shot approach using real anomaly samples from the test set. This comparison allows us to assess the validity of Gaussian noise as a substitute for true anomalies. Our results suggest that Gaussian noise perturbation provides competitive and in some cases even superior performance in detecting anomalies in different datasets, as shown in Figure~\ref{GA_eff}.
Substituting real anomaly samples with Gaussian noise led to slight changes in the AUROC score: MVTec decreased from 92.3 to 91.9, VisA from 93.4 to 93.2, MPDD from 94.7 to 94.3, while SDD increased from 93.22 to 94.1. ColonDB was excluded due to the absence of normal samples. These results suggest that Gaussian noise can improve performance by providing various deviations that generalize the anomaly representation better than specific, real anomalies — benefiting datasets with complex features such as MVTec, VisA and MPDD. However, for more uniform datasets such as SDD, real anomalies provided better results as they closely matched the structure of the dataset. Gaussian noise is thus a robust alternative for diverse anomalies, while real anomalies are preferable for simpler datasets.

%% file: 10_conclusion.tex
\section{Conclusion}
\label{sec:conclusion}
In this paper, we present Adaptive Prompt Tuning (APT), a novel few-shot framework for VLM-based anomaly detection that requires no prior knowledge and overcomes the limitations of human-designed prompts. Leveraging our proposed Self-Optimizing Meta-Prompt Guiding Scheme (SMGS), APT effectively optimizes learnable prompts to accurately capture context-specific anomalies, eliminating the need for labeled anomaly data or pre-existing anomaly semantics. Extensive experimental results demonstrate substantial performance improvements over traditional manually crafted prompts, confirming the robustness and effectiveness of our method in achieving precise, context-adaptive pixel-level anomaly detection.
While our current study primarily addresses pixel-wise anomaly detection, future research will focus on extending the capabilities of APT to class-level anomaly detection. This advancement will further enhance anomaly detection performance, offering greater versatility and applicability across various anomaly detection tasks and broader granularity levels.

\label{sec:conclusion}